\documentclass[10pt,twocolumn,letterpaper]{article}

\usepackage[pagenumbers]{cvpr} 
\usepackage{adjustbox}

\usepackage{algpseudocode}
\usepackage{algorithm}

\usepackage{subcaption}
\usepackage{booktabs} 

\usepackage{multirow}
\usepackage[table,xcdraw]{xcolor}

\newtheorem{theorem}{Theorem}[section]

\newtheorem{definition}{Definition}[section]

\newtheorem{observation}[theorem]{Observation}

\definecolor{cvprblue}{rgb}{0.21,0.49,0.74}
\usepackage[pagebackref,breaklinks,colorlinks,allcolors=cvprblue]{hyperref}

\def\paperID{6632} 
\def\confName{CVPR}
\def\confYear{2026}

\title{Fewer Tokens, Greater Scaling: Self-Adaptive Visual Bases \\ for Efficient and Expansive Representation Learning }

\author{Shawn Young, Xingyu Zeng, Lijian Xu*\\
Faculty of Computer Science and Control Engineering\\
Shenzhen University of Advanced Technology\\
Shenzhen, China
}

\begin{document}
\maketitle

\begin{abstract}

This paper investigates the fundamental relationship between model capacity and the minimal number of visual tokens required to preserve image semantics. Inspired by the Minimum Description Length principle, we reinterpret image tokens as vectors in a visual semantic space and define the intrinsic semantic complexity of an image as the smallest set of basis vectors needed to span this space. Building on this perspective, we propose Orthogonal Filtering, a lightweight module that adaptively clusters redundant tokens into a compact set of orthogonal bases. Through extensive experiments across a range of ViT models, we reveal a consistent token–model scaling law: larger models require significantly fewer tokens to span visual semantic space. Besides, we also contribute a visual long-context dataset.



\end{abstract}

\section{Introduction}
\label{sec:intro}

Despite the remarkable success of transformers in large language models, scaling up visual models faces a fundamental challenge: the inherent redundancy in images \cite{yangMaskedImageContrastive2024}. Unlike textual tokens, which are discrete and information-dense, visual tokens carry only limited semantic content as verified by MAE\cite{he2022masked}, where only a basic understanding of objects and scenes is needed to predict missing patches from their surroundings. As a result, sparse semantics in visual tokens pose a key challenge for scaling visual transformers efficiently.

While recent efforts in vision transformers attempt to mitigate redundancy through token pruning, clustering, or dynamic selection, these approaches typically operate within the feature space of a specific model and rely on heuristic importance measures. Such strategies often overlook the intrinsic semantic structure of visual data, and consequently, their effectiveness heavily depends on model architecture and training objectives. More importantly, despite the widespread recognition of redundancy in natural images, there remains a lack of quantitative understanding of how image redundancy fundamentally constrains the scalability of visual models.

\begin{figure}[htbp]
    \centering
    \begin{subfigure}[t]{0.48\textwidth}
        \centering
        \includegraphics[width=0.99\textwidth]{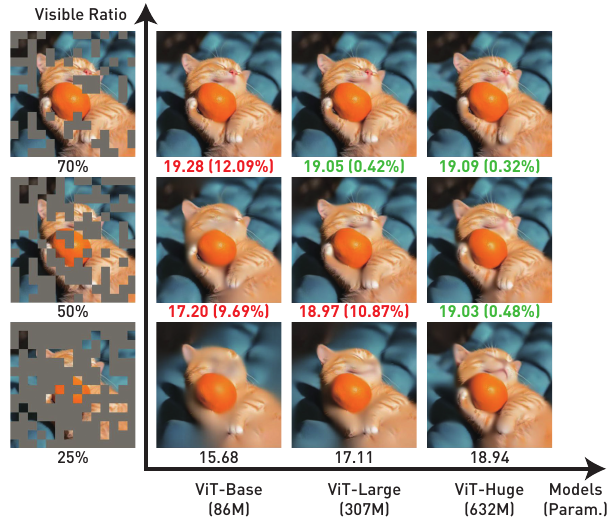}
        \caption{}
        \label{fig:cases-img}
    \end{subfigure}

    \begin{subfigure}[t]{0.48\textwidth}
        \centering
        \includegraphics[width=0.99\textwidth]{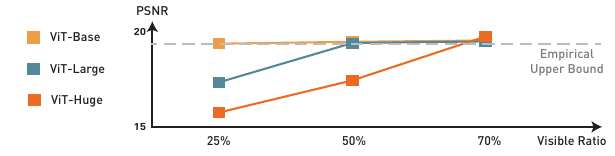}
        \caption{}
        \label{fig:cases-bound}
    \end{subfigure}
    \caption{\textbf{Larger models reach their performance upper bound with fewer tokens.} (a) Reconstruction results across different visible ratios and model capacities. PSNR reflects the performance at each visible ratio for a fixed model size, where red indicates improvement compared to the previous ratio and green denotes stability. (b) Tokens required for models of different capacities to reach the upper performance bound. An empirical performance upper bound is delineated with a grey line.
    }
    \label{fig:cases}
\end{figure}






The Minimum Description Length (MDL) theory \cite{rissanen1978modeling, grunwald2007minimum, zhao2023large} provides a new perspective to quantify semantic redundancy among visual tokens, interpreting the compression gain of each token as a measure of its semantic contribution. We posit that tokens are treated as vectors in a visual semantic space, and the objective becomes identifying a minimum set of base vectors that can effectively span this space. Consequently, under the lens of MDL, the semantic complexity of an image can be characterized by the minimal collection of bases sufficient to encode its essential visual information.

Within the MDL framework, we revisit the relationship between the number of image tokens and the model scaling, leveraging the reconstruction task of MAE pre-training \cite{he2022masked} as exhibited in Fig.\ref{fig:cases}. The aim is to determine the minimal number of token bases required for models of different scales to effectively span the semantic space of the original image. Our analysis reveals critical findings:

\begin{observation}
\label{ob.relation}


Specifically, beyond findings in MAE that larger models achieve better performance for the same visible ratio, we further empirically observe that as model scaling increases, fewer tokens are required to reach the performance upper bound. It highlights a complementary relationship between model capacity and token budget: larger models can achieve superior performance with fewer tokens, paving the way for the training and inference of larger vision models.  
\end{observation}

\begin{observation}
\label{ob.base}
Moreover, even the ViT-Base model with relatively few parameters can be trained using only 70\% of the tokens, revealing the substantial redundancy present in visual representations.
\end{observation}

\begin{observation}
\label{ob.generalization}
Besides achieving comparable performance on specific-tasks training, we observe that larger models exhibit stronger generalization ability, even when trained with fewer tokens. This observation highlights the significance of further exploring the intrinsic relationship between token utilization and model scaling.
\end{observation}

Therefore, summarizing the above findings of the qualitative relationship between model capacity and visible token ratio, under the case study of MAE, it raises the important question:


\textbf{\textit{What is the minimum set of image base vectors required to span the visual semantic space with larger models?}}

Inspired by mixture of experts (MoE) \cite{riquelme2021scaling,yang2025adapting}, we propose an orthogonal filter, a simple and lightweight module that self-adaptively clusters image tokens to construct a set of basis vectors spanning the visual semantic space.


We construct an orthogonal Filtering module composed of an allocator and multiple slots. The allocator clusters image tokens and assigns each token to a unique slot. Each slot fuses the assigned tokens to form a vector basis, while empty slots receive small random noise as the complement bases in the visual semantic space. The allocator is trained with an orthogonality loss to encourage semantic independence among slots.

Subsequently, a systematic exploration is conducted to reveal the scaling trend between the number of image tokens and model capacity. By progressively varying the aggregated tokens spanning different model scales, we empirically investigate the boundary of visual redundancy across various model capacities. Leveraging this empirical scaling trend, substantial reductions is achieved in training and inference cost spanning various vision models, while remaining fully plug-and-play and compatible with existing models.

Finally, inspired by Deepseek-OCR \cite{wei2025deepseekocrcontextsopticalcompression}, we hypothesize that once visual models overcome the bottleneck of visual redundancy, they may replicate the success trajectory of large language models. To facilitate this direction, we construct and release a visual long-context dataset, consisting of 17,365 papers collected from ICLR 2024 and 2025. Each paper is converted into a high-resolution long image of size 1275×16,500, providing a valuable resource for studying visual redundancy and long-context understanding in vision models.



In summary, our paper mainly makes the following contributions:
\begin{enumerate} 
    \item First, we establish a novel theoretical framework that formalizes visual tokens as vectors in a visual semantic space through the lens of minimum description length theory, enabling systematic analysis of the relationship between model capacity and visible token ratio from the perspective of the image instead of the model.


    \item Second, we propose an orthogonal Filtering module that adaptively clusters and merges image tokens into a compact set of orthogonal bases. Based on this design, we conduct a systematic exploration of the relationship between token redundancy and model capacity, revealing an empirical scaling trend and the boundary of visual tokens sustaining complete semantics.


    \item Leveraging this empirical trend, we demonstrate that our module substantially reduces training and inference cost spanning various vision models, while remaining plug-and-play and compatible with existing checkpoints. Notably, it further shows strong generalization in larger models, paving the way toward scalable visual intelligence from the perspective of image redundancy.  

    \item Finally, we contribute a visual long-context dataset, consisting of 17,365 high-resolution long image of papers collected from ICLR 2024 and 2025. It provides a valuable resource for studying visual redundancy and long-context understanding in vision models.
    
\end{enumerate}

Besides, we provide more related works in Appendix.






\section{Methodology}
\label{sec:method}

\begin{figure*}[htbp]
    \centering
    \begin{subfigure}[t]{0.8\textwidth}
        \centering
        \includegraphics[height=0.28\textheight]{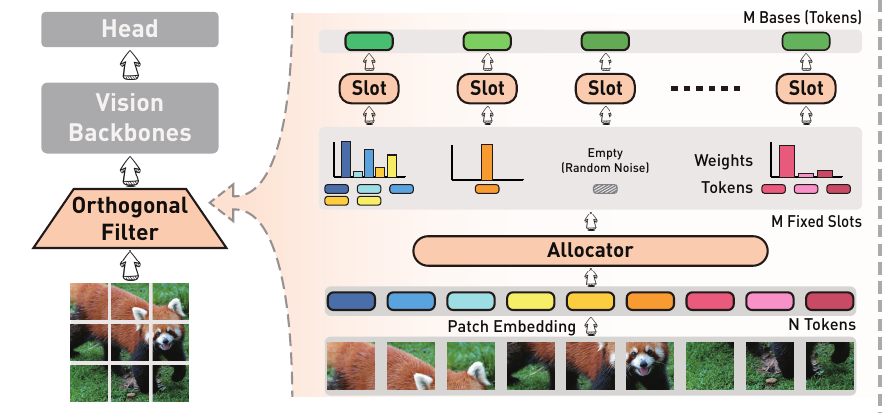}
        \caption{Orthogonal Bases for Representation Learning}
        \label{fig:slots}
    \end{subfigure}
    \begin{subfigure}[t]{0.18\textwidth}
        \centering
        \includegraphics[height=0.28\textheight]{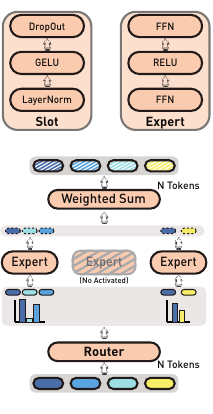}
        \caption{Structural Differences}
        \label{fig:moe}
    \end{subfigure}
    \caption{
\textbf{(a) A simple and lightweight orthogonal filter module} precedes the visual backbone to construct orthogonal bases for visual representation. Each image token is allocated to a unique slot where tokens in the same slot are weighted fused, while empty slots are filled with random noise. The allocator is responsible for extracting the visual bases guided by the orthogonality loss, while the slots merely normalize and fuse the assigned image tokens without performing additional feature extraction.
\textbf{(b) Structural Differences with MoE:} Unlike MoE, where tokens are distributed across multiple experts, our allocator groups semantically similar tokens into slots that collectively form the orthogonal bases. Consequently, MoE may contain inactive experts, whereas our design substitutes missing tokens with random noise. In addition, the number of output tokens corresponds to the number of slots rather than the input tokens, and our slots omit FFN layers used by MoE experts for feature extraction.    
    }
    \label{fig:analysis}
\end{figure*}



\subsection{Sparse Factorization under the MDL}

In this section, we first extend the Minimum Description Length (MDL) theory to token compression,  highlighting the phenomenon wherein the semantic space of an image can be expressed by a set of visual bases. Given an image represented by $N$ visual tokens $X = [ x_{1}, x_{2}, \cdots, x_{N} ]^{T} \in \mathbb{R}^{N\times d}$, the set of visual bases spanning the whole semantic space is expressed as $B = [ b_{1}, b_{2}, \cdots, b_{M} ]^{T} \in \mathbb{R}^{m\times d}$, where $M\ll N$ denotes the number of visual bases, and $d$ symbolizes the dimension of tokens. Consequently, the allocater dynamically assigns each token to one of $M$ latent slots, where the assignment is represented by a matrix:
\begin{equation}
    A = [a_{1},\cdots,a_{N}]^{T} \in \mathbb{R}^{N\times M}
    \label{eq.assign}
\end{equation}
where
\begin{equation}
0\leq A_{ij} \leq 1, \mathbf{1}^{T}A_{i}=1    
\label{eq.constrain}
\end{equation}
denotes the contribution of the $j$-th slot to the $i$-th tokens. Accordingly, we formally define the aggregation in various slots that produces a low-rank approximate reconstruction of the visual semantic space, as follows:
\begin{equation}
\hat{X} = AB, \text{rank}(\hat{X}) \leq M
\label{eq.fac}
\end{equation}
It parallels low-rank factorization of the dense feature map \cite{hu2022lora}, but differs in that both the assignment matrix A and the basis matrix B are jointly parameterized and learned through gradient descent, with allocator sparsity and orthogonality constraints ensuring semantic disentanglement.

Therefore, we extend the Minimum Description Length (MDL) theory \cite{shalev-shwartzUnderstandingMachineLearning2014} into visual representation as follows:

\begin{theorem}
\label{theo.mdl}
Let $\mathcal{H}$ be a hypothesis class of visual representations induced by the factorization module, and $h = (A,B)\in \mathcal{H}$ denotes a specific instantiation characterized by the assignment matrix $A$ and the basis matrix $B$. Let $d: \mathcal{H}\rightarrow \{0,1\}^{*}$ represent a prefix-free description of $h$, and let $|d|$ symbolize the code length of $d(h)$. For any dataset of images drawn i.i.d. from a distribution $\mathcal{D}$ over the visual domain, define the empirical reconstruction loss and expected reconstruction loss respectively as:
\begin{equation}
    \begin{aligned}
        \mathcal{L}_{S}(h) = \frac{1}{m} \sum_{i=1}^{m} ||X_{i} - A_{i}B_{i} ||^{2} \\
        \mathcal{L}_{\mathcal{D}}(h) = \mathbb{E}_{X\sim D} ||X -AB || ^{2}
    \end{aligned}
\end{equation}
where $S\sim \mathcal{D}^{m}$ has probability at least $1-\delta$ over rhe sampled training set. Then for every confidence level $\delta > 0$, we have:
\begin{equation}
    \mathcal{L}_{\mathcal{D}}(h) \leq \mathcal{L}_{S}(h) + \sqrt{\frac{(|h|+\ln{(2/\delta)})}{2m}}
    \label{eq.mdl}
\end{equation}

\end{theorem}

Consequently, Eq.\ref{eq.mdl} provides a generalization bound for the visual compression model under the MDL principle. The expected reconstruction error $\mathcal{L}_{\mathcal{D}}(h) $ is upper-bounded by the empirical reconstruction loss $ \mathcal{L}_{S}(h)$ and a complexity penalty proportional to the description length $|h|$ of the representation parameters. Since the code length $|h|$ increases with the number of bases $M$ and the effective sparsity of $A$, minimizing the total description length corresponds to selecting the most compact set of bases that still achieves low reconstruction loss.

This bound justifies our findings in Observation \ref{ob.relation} and \ref{ob.generalization}, that the low-rank factorization naturally implements an MDL-consistent compression of the image semantics, where reducing redundant tokens with smaller $|h|$ leads to improved generalization as long as semantic fidelity $\mathcal{L}_{S}(h)$ remains low.

In essence, larger models admit shorter visual descriptions—they require fewer bases $M$ to reconstruct the same semantic space, providing a theoretical underpinning for the observed scaling trend between model capacity and the minimal token ratio.

\subsection{Allocator with Orthogonal Bases}
\label{sec.Allocator}

Building upon the Eq.\ref{eq.fac} and Eq.\ref{eq.mdl}, where image redundancy was modeled as a low-rank factorization constrained by the MDL principle, we now instantiate this formulation through a practical module—the Orthogonal Filtering Allocator.
While the previous section established that each image can be efficiently represented by a compact set of semantic bases $B$ learned via sparse assignments $A$ in Eq.\ref{eq.assign}, this subsection focuses on how these assignments are dynamically produced and optimized during training.
The key challenge lies in constructing a mechanism that can automatically cluster semantically similar tokens into shared slots, while ensuring that the resulting bases remain orthogonal, balanced, and information-preserving across different spatial regions.

\begin{algorithm}
\caption{Simple Orthogonal Filter Module}
\label{alg:moe}
\begin{algorithmic}[1]
\State \textbf{Input:} $x \in \mathbb{R}^{N \times d}$ 
\State \textbf{Output:} $y \in \mathbb{R}^{M \times d}$, $\mathcal{L}_{\text{orth}}$

\Statex
\State \textbf{Step 1: Gating and Allocating}
\State $A \gets \text{Softmax}(\text{Linear}(x))$ \Comment{Slot probabilities}
\State $I \gets \text{argmax}(A, \text{dim}=-1)$ \Comment{Selected slot indices}
\State $W \gets \text{gather}(A, I)$ \Comment{Routing weights}

\Statex
\State \textbf{Step 2: Slot Fusion}
\State $Y \gets \text{zeros}(M, d)$ \Comment{Initialize output tensor}
\For{$k = 1$ to $M$}
    \State $\text{mask} \gets (I = k)$ \Comment{Tokens assigned to slot $k$}
    \If{$\text{any}(\text{mask})$}
        \State $\text{tokens}\ E_{k} \gets x[\text{mask}]$, $\text{weights} \gets W[\text{mask}]$
        \State $\text{expert\_out} \gets \text{Expert}_e(\text{$E_{k}$}) \odot \text{weights}$
        \State $Y[k,:] \gets \text{scatter\_add}(\text{expert\_out})$
    \Else
        \State $Y[k,:] \gets \text{Expert}_e(\text{random}(d))$ \Comment{Handle empty expert}
    \EndIf
\EndFor

\Statex
\State \textbf{Step 3: Orthogonal Loss}

\If{$\text{is\_training}$}
    \State $\text{sim} \gets \text{cosine\_similarity}(x, Y)$  \Comment{Compute similarity between tokens and slot outputs}
    \State $L_{\text{orth}} \gets \text{OrthogonalLoss}(E_{k}, \text{sim}, \tau)$  \Comment{Compute orthogonal loss using Eq.\ref{eq.likelihood} and Eq.\ref{eq.orth_loss}}
\Else
     \State $L_{\text{orth}} \gets 0$ \;
\EndIf

\State \textbf{return} $Y, L_{\text{orth}}$
\end{algorithmic}
\end{algorithm}

Inspired by mixture of experts (MoE) \cite{riquelme2021scaling}, we observe that the routing mechanism naturally serves as an effective implementation of the assignment matrix $A$ in Eq.\ref{eq.assign}. Thus, we design the allocator, dynamically determining the token-to-slot allocation, satisfying the probabilistic constraint in Eq.\ref{eq.constrain}, thereby realizing the sparse factorization derived in our theoretical formulation of Eq.\ref{eq.fac}. Architecturally, unlike MoE as presented in Fig.\ref{fig:moe}, our allocator assigns each token to its most probable slot, thereby eliminating the need for the top-k gating mechanism used in MoE to select multiple experts. Furthermore, by grouping similar tokens into the same slot, our proposed slot differs from MoE's experts, which rely on FFN layers for feature extraction. Instead, our slots are primarily responsible for normalizing and aggregating information across tokens.

Fundamentally, in contrast to the router in MoE, which relies on an auxiliary loss function for load balancing across experts, our allocator is driven by an orthogonality loss designed to shape each slot into a set of basis vectors in the visual semantic space. It encourages tokens within the same slot to be semantically similar, while ensuring distinct slots capture divergent semantic concepts, which are orthogonal in the visual representation space.

Formally, $E_{k} = \{ x_{i}: \mathop{\arg\max}\limits_{j}\ \  \text{sim}(x_{i},b_{j}) = k \}$ denotes the set of image tokens assigned to the $k$-th slot, where $\text{sim}$ represents the similarity measurement, and we use the cosine similarity here. Tokens from the same slot $x_{i}\in E_{k}$ are treated as positive pairs with their generated basis $b_{k}$,  while tokens from different slots are treated as negative pairs with other image bases. We concurrently maximise the similarity of $|E_{k}|$ positive pairs while minimizing the similarity of negative examples to drive the allocator. Thus, the log likelihood of each slot $E_{k}$ follows:

\begin{equation}
\label{eq.likelihood}
\begin{aligned}
    &l(E_{k}, B)  = \\
    & 
    \frac{1}{|E_{k}|}\sum_{x_{i}\in E_{k}} 
    \log
    \frac
    {\exp(\text{sim}(x_{i},b_{k})/\tau)}
    {\sum_{j=1}^{M}
        \mathbb{1}_{ \left[ j \neq k \right] } \exp(\text{sim}(x_{i},b_{j})/\tau)
    }    
\end{aligned}
\end{equation}
where $\mathbb{1}_{ \left[ j \neq k \right] } \in \{0, 1\}$ denotes an indicator function evaluating to 1 if $j\neq k$, and $\tau$ is a trainable temperature parameter to effectively scale the different samples. Consequently, we propose the orthogonal loss to train the allocator for low-rank factorization as follows:

\begin{equation}
\label{eq.orth_loss}
    \mathcal{L}_{orth} = 
     -
    \frac{1}{M} \sum_{k=1}^{M}
    \mathbb{E}_{M}\left[
    l(E_{k}, B) 
    \right]
\end{equation}


Thus, through the orthogonal loss, we enable the learning of orthogonal filters that form a set of basis functions to represent the visual semantic space. Moreover, this module is straightforward to implement, as exhibited in Algorithm \ref{alg:moe}. Designed for seamless integration, this plug-and-play module can be inserted at any network layer to perform token fusion.

\subsection{The Law of Parametric Efficiency Priority}

Obtaining the orthogonal filter in Section \ref{sec.Allocator}, our focus toward empirically investigating \textit{What is the minimum set of image base vectors required to span the visual semantic space with larger models?} To this end, we place the orthogonal filter solely at the front end of the vision backbone, where it aggregates image tokens into a compact set of orthogonal bases representing the image.

In this section, we start from the general case and empirically verify that, given the same number of image bases, models with different parameter scales exhibit performance trajectories consistent with the scaling law. We further fix the model architecture to investigate the quantitative relationship between the number of image bases and its attainable performance, through which we derive the empirical Minimum Description Length (MDL) corresponding to each model size. By comparing these empirical MDLs across models of varying capacities, we identify an efficiency frontier that characterizes the trade-off between model capacity and image redundancy. This leads us to formulate The Law of Parametric Efficiency Priority, which states that larger models achieve the upper bound of visual performance with fewer image tokens—providing a theoretical basis and practical guidance for efficient training and scaling of large vision models.

\begin{table}[htbp]
    \centering
    \begin{adjustbox}{width=.48\textwidth}{
    \begin{tabular}{@{}cccccc@{}}
    \multicolumn{6}{c}{
    \includegraphics[width=0.6\textwidth]{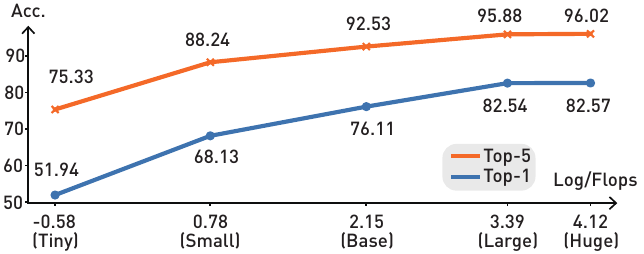}
    }                                                                  \\
    \\
    \toprule
    Models   & ViT-Tiny & ViT-Small      & ViT-Base      & ViT-Large      & ViT-Huge       \\ \midrule
    Flops/G  & 0.56 	& 2.19           & 8.62 	     & 29.78 	      & 61.76  \\
    Param./M & 12.85 	& 50.47          & 200.03        & 505.94 	      & 946.91 \\ \bottomrule
    \end{tabular}}\end{adjustbox}
    \caption{\textbf{The scaling law of various ViT models under the same number of 96 image bases.} Top-1 accuracy is applied to evaluate the performance of spanning visual semantic space. ImageNet is leveraged for training and validating, with the resolution of 224$\times$224. Besides, we also provide FLOPs/G and Parameters (Param.)/M to assess the performance of various ViT models under 96 image bases.
    }
    \label{tab:img-bases}
\end{table}

Inspired by the scaling law \cite{kaplan2020scaling}, we observe that the scaling law also manifests in the visual domain, particularly when the number of image bases is fixed. As illustrated in Table \ref{tab:img-bases}, we plot model performance against the logarithm of Floating Point Operations Per Second (FLOPs) across models of increasing capacity. The results reveal that, before reaching the task-specific performance ceiling, the model performance follows a clear power-law relationship with respect to model size under a fixed token budget. This finding highlights a critical implication: when the number of tokens remains constant, achieving further performance gains requires an exponentially increasing computational cost.

\begin{table}[htbp]
    \centering
    \begin{adjustbox}{width=.48\textwidth}{
    \begin{tabular}{@{}ccccccc@{}}
    \multicolumn{7}{c}{
    \includegraphics[width=0.6\textwidth]{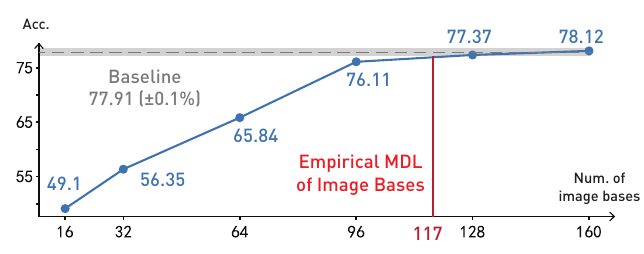}
    }                                                                  \\ \toprule
    Slots   & 16      & 32      & 64      & 96      & 128     & 160    \\ \midrule
    Flops/G & 1.72    & 3.10   & 5.87   & 8.62   & 11.39  & 14.15 \\
    Param./M & 105.35   & 124.29   & 162.16   & 200.03   & 237.90  & 275.78 \\ \bottomrule
    \end{tabular}}\end{adjustbox}
    \caption{\textbf{The empirical MDL of image bases under ViT-Base.} Top-1 accuracy is applied to evaluate the performance of spanning visual semantic space with different image bases. ImageNet is leveraged for training and validating, with the resolution of 224$\times$224. The baseline denotes the original results of ViT-Base. Besides, we also provide FLOPs/G and Parameters (Param.)/M to assess the performance of ViT-base model under various numbers of image bases.
    }
    \label{fig:vit-slots}
\end{table}


Furthermore, as shown in Table \ref{tab:img-bases}, the results of ViT-Huge indicate that once the model reaches its performance upper bound, it no longer requires as many image bases as ViT-Large. It is worth noting that this upper bound is jointly determined by the model’s parameter scale, the training dataset, and the complexity of the task. Motivated by this finding, we further analyze the effect of varying the number of image bases under the same model configuration, as illustrated in Table \ref{fig:vit-slots}. Remarkably, even with only 16 image bases, the model achieves strong performance, attaining a Top-1 accuracy of 49.1, which empirically verifies our earlier Observation \ref{ob.base}. As the number of image bases gradually increases, the relationship between model performance and image bases also follows a power-law trend until it saturates at the performance ceiling. Based on this power-law curve, we empirically infer the Minimum Description Length (MDL) of image bases for each model. Interestingly, the required MDL is substantially smaller than the number of tokens used in standard vision models, such as 117 MDL of ViT-Base, which is far fewer than the original 196 input tokens.

\begin{table}[htbp]
    \centering
    \begin{adjustbox}{width=.48\textwidth}{
    \begin{tabular}{@{}cccccc@{}}
    \multicolumn{6}{c}{
    \includegraphics[width=0.6\textwidth]{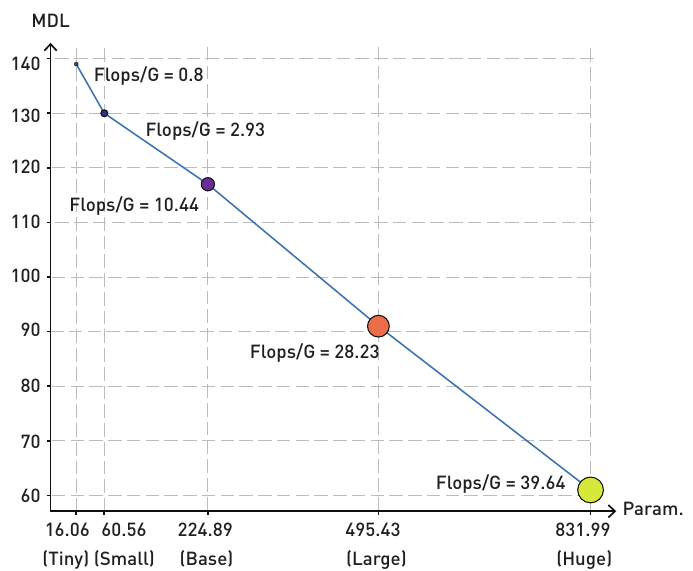}
    }                                                                  \\
    \\
    \toprule
    Models  & ViT-Tiny & ViT-Small & ViT-Base & ViT-Large & ViT-Huge \\ \midrule
    MDL	    & 139  	   & 130       & 117      & 91        & 61       \\
    Flops/G	& 0.80 	   & 2.93      & 10.44    & 28.23     & 39.64    \\ 
    Param/M	& 16.06    & 60.56     & 224.89   & 495.43    & 831.99   \\ \bottomrule
    \end{tabular}}\end{adjustbox}
    \caption{\textbf{The Law of Parametric Efficiency Priority.} We provide MDL, FLOPs/G and Parameters (Param.)/M among various scaling ViT models. ImageNet is leveraged for training and validating, with the resolution of 224$\times$224.}
    \label{tab.pep}
\end{table}


Consequently, obtaining the empirical MDL for each model, we further observe a clear relationship between the MDL and the model’s parameter scale, summarized as the Law of Parametric Efficiency Priority (LPEP), presented in Table \ref{tab.pep}. Specifically, as the number of vision model parameters increases, the required MDL decreases approximately linearly. This implies that larger models require proportionally fewer image bases to achieve optimal performance, resulting in only a linear growth in computational cost (FLOPs) for increasingly powerful vision models, rather than the exponential growth exhibited in Table \ref{tab:img-bases}. Formally, We define the Law of Parametric Efficiency Priority (LPEP) under MDL theory:

\begin{definition}
For a family of models trained on a data distribution from a sufficiently complex domain, the number of tokens required for a model to approach its irreducible performance ceiling scales down as the number of non-embedding parameters increases. Formally, a power-law relationship can be used to model our proposed law of PEP:
\begin{equation}
    \eta_{B}(\theta) \propto \frac{C}{\theta^{\alpha}}
\end{equation}
where $\eta_{B}(\theta)$ denotes the critical number of image bases $B$ required for a model with $\theta$ parameters to saturate its performance upper bound, $C$ is a constant that represents the intrinsic "complexity" of the visual semantic space, and $\alpha$  determines how strongly an increase in parameters reduces the bases token requirement as a positive constant $0<\alpha<1$.
\end{definition}

Therefore, the law of Parametric Efficiency Priority provides an empirical quantitative guideline for scaling vision models: larger models can be trained with substantially fewer tokens while preserving full semantic capacity. By revealing the functional relationship between model size and the minimal token budget required for semantic completeness, this law offers a principled foundation for future research on efficient training strategies for large-scale vision models.

\subsection{Visual Long-context Dataset for Scaling Models}

As pioneers in exploring the relationship between the MDL of image bases and model capacity, we deeply recognize the scarcity of large-scale visual long-context datasets, particularly those suitable for training high-capacity vision models. In particular, inspired by DeepSeek-OCR \cite{wei2025deepseekocrcontextsopticalcompression}, we envision that vision models can achieve autonomous reasoning through scalable expansion, replicating the remarkable success of large models in the language domain while alleviating the computational bottlenecks commonly encountered in LLMs.

In this context, a more realistic training dataset for large vision models is required to validate their potential to be trained under the visual long-context. Recognizing the demand for higher-quality visual long-context images, we develop a dataset called PaperScope. This novel dataset consists of 17,365 high-resolution long-context images of papers collected from ICLR 2024 and 2025. It provides a valuable resource for studying visual redundancy and long-context understanding in vision models.
More details are given in Appendix.

\section{Experiments}

In this section, we evaluate the behavior of vision models under the Minimum Description Length perspective, with a particular focus on their generalization ability and robustness across downstream tasks. Our experiments span domain generalization, object detection, and a suite of out-of-distribution benchmarks, aiming to verify Observation \ref{ob.generalization}, that larger models not only require fewer tokens to reach their performance ceiling, but also exhibit superior generalization and robustness. These findings collectively pave the way toward more scalable and efficient large-scale vision models. And more details are provided in the Supplementary Material.

\subsection{Generalizion with Larger Models}


First, we conduct experiments of domain generalization setting following the SiameseIM \cite{taoSiameseImageModeling2023}, verifying the various models under MDL as presented in Table \ref{table.gen}.

\begin{table}[htbp]
    \centering
    \begin{tabular}{@{}lccccc@{}}
    \toprule
    Methods                                           & Flops/G & IN-A & IN-R & IN-S      \\ \midrule
    MSN       \cite{assranMaskedSiameseNetworks2022}  &  17.4   & 37.5 & 50.0 & 36.3      \\
    iBOT      \cite{zhouImageBERTPretraining2021a}    &  17.5   & 42.4 & 50.9 & 36.9      \\
    DenseCL   \cite{wangDenseContrastiveLearning2021} &  74.7   & 30.8 & 43.8 & 29.9      \\
    MAE       \cite{heMaskedAutoencodersAre2022}      &  17.5   & 35.9 & 48.3 & 34.5      \\
    OCL       \cite{yangMaskedImageContrastive2024}   &  12.0   & 42.2 & 52.3 & 37.6      \\ \midrule
    \rowcolor[HTML]{EFEFEF} 
    Ours/ViT-B@117                                    &  10.4   & 42.5 & 52.7 & 38.3      \\
    \rowcolor[HTML]{EFEFEF} 
    Ours/ViT-L@91                                     & 28.2    & 59.5 & 60.1 & 44.9      \\
    \rowcolor[HTML]{EFEFEF}
    Ours/ViT-H@61                                     &  39.6   & 68.3 & 65.1 & 49.3      \\
    \bottomrule
   \end{tabular}
    \caption{\textbf{Comparison with previous methods on generalization capability and robustness on ImageNet-A (IN-A) \cite{hendrycksNaturalAdversarialExamples2021}, ImageNet-R (IN-R) \cite{hendrycksManyFacesRobustness2021} and ImageNet-Sketch (IN-S) \cite{wangLearningRobustGlobal2019} datasets.}  We provide the results of ViT-base model with MDL of 117 image bases and the ViT-Huge model with MDL of 61. Top-1 accuracy is used as the metric. The resolution of images is fixed to 224$\times$224. FLOPs/G is utilized to show the runtime and computational resources of the training.
    }
    \label{table.gen}
\end{table}

As exhibited in Table \ref{table.gen}, our method achieves a significantly more favorable efficiency–performance trade-off than existing approaches. Ours/ViT-B@117 attains performance on ImageNet-A comparable to or slightly better than competitive methods such as MSN and iBOT, while reducing FLOPs by more than 40\%. This indicates that the compact visual bases learned by our Orthogonal Filtering framework preserve essential semantic information even under extreme domain shift, allowing generalized and robust representation learning without relying on heavy computational budgets. 

Besides, a striking observation emerged from our generalization study: the larger models, even with fewer image bases, are improving the robustness. With only 61 visual bases, Ours/ViT-H@61 outperforms all baselines and smaller models by a large margin on all three benchmarks. It reveals a clear empirical scaling trend, that larger models exhibit higher token efficiency, meaning they can extract semantics from fewer input tokens while producing more generalized and stable representations. Such behavior aligns with our proposed Law of Parametric Efficiency Priority and provides strong empirical evidence that model capacity compensates for reduced input resolution in the token space, enabling more efficient visual learning at scale.

Moreover, the consistent improvements observed across ViT-Base, ViT-Large, and ViT-Huge demonstrate that our approach generalizes well across different model capacities and does not rely on architectural peculiarity. 

\subsection{Robustness within Fewer Tokens}

Furthermore, we conduct experiments of out-of-distribution (OOD) to validate the robustness of our methods with fewer tokens.

\begin{table}[htbp]
    \centering
    \begin{tabular}{@{}lcccc@{}}
    \toprule
    \multirow{2}{*}{Methods}                                   & \multicolumn{2}{c}{SVHN}                                   & \multicolumn{2}{c}{LSUN}                                       \\ \cmidrule(l){2-5}                 
                                                               & FPR                       & AUROC                          & FPR                             & AUROC                        \\ \midrule                             
    ProxyAnchor \cite{kimProxyAnchorLoss2020}                  & 87.2                      & 82.4                          & 37.2                            & 91.7                          \\ 
    CE + SimCLR \cite{winkensContrastiveTrainingImproved2020}  & 24.8                      & 94.5                          & 56.4                            & 89.0                          \\ 
    SSD+        \cite{sehwagSSDUnifiedFramework2020}           & 31.2                      & 94.2                          & 79.4                            & 85.2                          \\ 
    CIDER       \cite{mingHowExploitHyperspherical2023}        & 12.6                      & 97.8                          & 30.2                            & 92.8                          \\ \midrule
    \rowcolor[HTML]{EFEFEF} 
    Ours/ViT-B@117                & 13.0  & 95.1  & 33.7     & 92.2   \\ \bottomrule       
    \end{tabular}
    \caption{Evaluation results of OOD detection with the OOD datasets of SVHN\cite{netzer2011reading} and LSUN \cite{yu2015lsun}. FPR$\downarrow$ and AUROC$\uparrow$ are applied to evaluate the performance of different methods.}
    \label{table:OOD-1}
\end{table}

Across both SVHN and LSUN OOD benchmarks, our method consistently achieves the second-best compared to the specialized OOD method CIDER, demonstrating strong robustness against distribution shifts. In particular, while several baselines exhibit large performance fluctuations across datasets, such as ProxyAnchor\cite{kimProxyAnchorLoss2020} and SSD+\cite{sehwagSSDUnifiedFramework2020} which degrade sharply on SVHN, our results remain stable. It indicates that the proposed method does not overfit to a specific distribution and generalizes reliably across substantially different OOD scenarios.

\subsection{Scaling for Various Tasks}

Beyond classification, we further aim to assess the scaling capability of our model, particularly its ability to transfer effectively across a diverse range of vision tasks, paving the way for large vision models. We evaluate various methods on COCO object detection and instance segmentation within Mask R-CNN framework \cite{heMaskRCNN2017}, where our ViT variants are used as backbones. Following the protocol of MixMAE \cite{liuMixMAEMixedMasked2023}, we adopt a 16$\times$16 window configuration to accommodate the 1024$\times$1024 input resolution.

\begin{table}[htbp]
    \centering
    \begin{tabular}{@{}lccc@{}}
    \toprule
    \multicolumn{1}{c}{\multirow{2}{*}{Methods}}                   & \multicolumn{2}{c}{COCO} & ADE20k \\
    \multicolumn{1}{c}{}                                           & $AP^{b}$   & $AP^{m}$    & mIoU   \\ \midrule
    \multicolumn{4}{l}{\textbf{ViT-B/16}}                                                              \\ 
    ViT   \cite{taoSiameseImageModeling2023}                & 47.9       & 42.9        & 47.4   \\
    BEiT         \cite{baoBEiTBERTPreTraining2021}                 & 49.8       & 44.4        & 47.1   \\
    MAE          \cite{heMaskedAutoencodersAre2022}                & 50.6       & 45.1        & 45.0   \\
    MixMAE       \cite{liuMixMAEMixedMasked2023}                   & 52.7       & 47.0        & 51.1   \\
    OCL          \cite{yangMaskedImageContrastive2024}             & 51.5       & 45.5        & 46.1   \\ 
    \rowcolor[HTML]{EFEFEF} 
    Ours/ViT-B@117  & 50.1 & 45.0 & 47.4 \\ \midrule
    \multicolumn{4}{l}{\textbf{ViT-L/16}}                                                              \\ 
    BEiT         \cite{baoBEiTBERTPreTraining2021}                 & 53.3       & 47.1        & 53.3   \\
    MAE          \cite{heMaskedAutoencodersAre2022}                & 53.3       & 47.2        & 53.6   \\
    MixMAE       \cite{liuMixMAEMixedMasked2023}                   & 54.3       & 48.2        & 53.8   \\ 
    OCL                                                            & 53.2       & 47.0        & 53.2   \\ 
    \rowcolor[HTML]{EFEFEF} 
    Ours/ViT-L@91    & 53.8 & 48.0 & 54.0                                                                                          \\ \bottomrule
    \end{tabular}
    \caption{\textbf{Comparison with previous methods on various downstream tasks, including object detection and segmentation on COCO and ADK20K.} We report $\text{AP}^{\text{box}}$ ($\text{AP}^{\text{b}}$) and $\text{AP}^{\text{mask}}$ ($\text{AP}^{\text{m}}$) on COCO, and mIoU on ADE20K. Arch. represents the model architecture, where ViT-B/16 and ViT-L/16 are utilized to validate the performance of various methods.
    }
    \label{table:coco}
\end{table}

From the ViT-Large@91 results, we observe that even when trained with only 46\% number of the whole tokens, our method maintains highly competitive performance. It achieves the best segmentation score on ADE20K and ranks second on COCO object detection, trailing MixMAE by merely 0.4\% in $AP^{m}$. These results indicate that removing redundant visual tokens does not hinder transferability. Instead, it effectively mitigates the overfitting tendencies typically observed in larger models.

Moreover, as the model scales up, the advantages of our approach become increasingly pronounced. Specifically, when compared with the ViT-B@117 results, our ViT-B within MDL of image bases is not the top-performing method among its peers. However, our ViT-L@91 achieves the best performance on COCO $AP^{b}$ and ADE20K mIoU, and trails the leading MixMAE  \cite{liuMixMAEMixedMasked2023}   by only 0.2 points in COCO $AP^{m}$. This clearly demonstrates that our method benefits substantially from model scaling. Notably, our scaling trajectory requires far fewer tokens and introduces only a 17.8G FLOPs increase, whereas conventional scaling from ViT-B to ViT-L typically incurs an additional 42.6G FLOPs, representing a 58\% reduction in computational overhead.

\subsection{Ablation Studies}



\subsubsection{Orthogonal Bases}


First, we examine whether the proposed orthogonal filter effectively clusters semantically similar tokens while maintaining sufficient separability across distinct visual bases. Using a ViT-B@117 model, tokens mapped to the same slot are treated as belonging to the same class, whereas tokens allocated to different slots are regarded as different classes. Based on cosine similarity, we compute intra-class compactness and inter-class separability.

Our results show that the intra-class compactness reaches 0.1843, demonstrating that tokens grouped into the same slot exhibit high semantic consistency. In contrast, the inter-class separability achieves 1.8461, indicating substantial separation across slots. This strong inter-slots divergence suggests that the learned bases are nearly orthogonal, aligning well with our low-rank decomposition approximation in Eq.\ref{eq.fac} and supporting that generated image bases collectively span the underlying visual semantic space.

\subsubsection{Larger Images with High resolution}


\begin{table}[htbp]
\centering
\begin{tabular}{@{}ccccc@{}}
\toprule
Resolution & 224  & 384  & 512  & 1024  \\
Patches    & 196  & 576  & 1024 & 4096  \\ \midrule
Accuracy   & 82.5 & 82.9 & 83.6 & 74.31 \\ \bottomrule
\end{tabular}
\caption{\textbf{Ablation experiments on image resolution within ViT-L@91 on ImageNet.} Patches are also provided.}
\label{tab:ab}
\end{table}

Subsequently, we further examine the influences of resolution on visual models under the MDL framework. Table \ref{tab:ab} reports the impact of varying resolution on ViT-L@91. We observe that the model consistently improves as the image resolution increases from 224 to 512, reflecting its ability to exploit the additional visual detail available at higher pixel densities. With more pixels per patch, the orthogonal bases can capture richer semantic structures while simultaneously filtering out redundant or visually similar content.

However, when it further increases to 1024, performance drops sharply. We attribute this decline to the mismatch between extremely fine-grained patches with the patch size of 16 and the limited 91 MDL image bases, exceeding the representational capacity of the 91 MDL image bases.

\section{Conclusion and Outlooks}


This paper introduces an MDL-driven perspective for understanding image redundancy and proposes Orthogonal Filtering to derive compact visual bases that preserve the semantic structure of the image. Through systematic scaling experiments, we reveal a consistent trend—larger models require fewer tokens to reach their performance ceiling—formulated as the Law of Parametric Efficiency Priority. Our module is plug-and-play, architecture-agnostic, and yields notable gains in efficiency, generalization, and robustness across diverse vision tasks.

Looking forward, we hope this study can facilitate the training of large vision models, where increasingly powerful architectures can be trained with progressively fewer image tokens. Furthermore, we plan to extend this line of research to long-context visual inputs, particularly exploring varying image resolutions.

{
    \small
    \bibliographystyle{ieeenat_fullname}
    \bibliography{main}

@String(ICCV= {Int. Conf. Comput. Vis.})

@String(ECCV= {Eur. Conf. Comput. Vis.})

@String(NIPS= {Adv. Neural Inform. Process. Syst.})

@String(ICLR = {Int. Conf. Learn. Represent.})

@String(AAAI = {AAAI})

@String(ICCV  = {ICCV})

@String(ECCV  = {ECCV})

@String(NIPS  = {NeurIPS})

@String(ICLR  = {ICLR})

@book{grunwald2007minimum,
  title={The minimum description length principle},
  author={Gr{\"u}nwald, Peter D},
  year={2007},
  publisher={MIT press}
}

@article{rissanen1978modeling,
  title={Modeling by shortest data description},
  author={Rissanen, Jorma},
  journal={Automatica},
  volume={14},
  number={5},
  pages={465--471},
  year={1978},
  publisher={Elsevier}
}

@article{zhao2023large,
  title={Large language models as commonsense knowledge for large-scale task planning},
  author={Zhao, Zirui and Lee, Wee Sun and Hsu, David},
  journal={Advances in neural information processing systems},
  volume={36},
  pages={31967--31987},
  year={2023}
}

@inproceedings{he2022masked,
  title={Masked autoencoders are scalable vision learners},
  author={He, Kaiming and Chen, Xinlei and Xie, Saining and Li, Yanghao and Doll{\'a}r, Piotr and Girshick, Ross},
  booktitle={Proceedings of the IEEE/CVF conference on computer vision and pattern recognition},
  pages={16000--16009},
  year={2022}
}

@article{riquelme2021scaling,
  title={Scaling vision with sparse mixture of experts},
  author={Riquelme, Carlos and Puigcerver, Joan and Mustafa, Basil and Neumann, Maxim and Jenatton, Rodolphe and Susano Pinto, Andr{\'e} and Keysers, Daniel and Houlsby, Neil},
  journal={Advances in Neural Information Processing Systems},
  volume={34},
  pages={8583--8595},
  year={2021}
}

@misc{wei2025deepseekocrcontextsopticalcompression,
      title={DeepSeek-OCR: Contexts Optical Compression}, 
      author={Haoran Wei and Yaofeng Sun and Yukun Li},
      year={2025},
      eprint={2510.18234},
      archivePrefix={arXiv},
      primaryClass={cs.CV},
      url={https://arxiv.org/abs/2510.18234}, 
}

@article{hu2022lora,
  title={Lora: Low-rank adaptation of large language models.},
  author={Hu, Edward J and Shen, Yelong and Wallis, Phillip and Allen-Zhu, Zeyuan and Li, Yuanzhi and Wang, Shean and Wang, Lu and Chen, Weizhu and others},
  journal={ICLR},
  volume={1},
  number={2},
  pages={3},
  year={2022}
}

@book{shalev-shwartzUnderstandingMachineLearning2014,
  title = {Understanding Machine Learning: {{From}} Theory to Algorithms},
  shorttitle = {Understanding Machine Learning},
  author = {{Shalev-Shwartz}, Shai and {Ben-David}, Shai},
  year = 2014,
  publisher = {Cambridge university press},
  urldate = {2025-10-28},
  langid = {american}
}

@inproceedings{assranMaskedSiameseNetworks2022,
  title = {Masked {{Siamese Networks}} for~{{Label-Efficient Learning}}},
  booktitle = {Computer {{Vision}} – {{ECCV}} 2022},
  author = {Assran, Mahmoud and Caron, Mathilde and Misra, Ishan and Bojanowski, Piotr and Bordes, Florian and Vincent, Pascal and Joulin, Armand and Rabbat, Mike and Ballas, Nicolas},
  editor = {Avidan, Shai and Brostow, Gabriel and Cissé, Moustapha and Farinella, Giovanni Maria and Hassner, Tal},
  year = {2022},
  pages = {456--473},
  publisher = {Springer Nature Switzerland},
  location = {Cham},
  doi = {10.1007/978-3-031-19821-2_26},
  isbn = {978-3-031-19821-2},
  langid = {english}
}

@inproceedings{zhouImageBERTPretraining2021a,
  title = {Image {{BERT Pre-training}} with {{Online Tokenizer}}},
  booktitle = {International {{Conference}} on {{Learning Representations}}},
  author = {Zhou, Jinghao and Wei, Chen and Wang, Huiyu and Shen, Wei and Xie, Cihang and Yuille, Alan and Kong, Tao},
  year = {2022},
  month = oct,
  urldate = {2024-11-11},
  langid = {english}
}

@inproceedings{wangDenseContrastiveLearning2021,
  title = {Dense {{Contrastive Learning}} for {{Self-Supervised Visual Pre-Training}}},
  author = {Wang, Xinlong and Zhang, Rufeng and Shen, Chunhua and Kong, Tao and Li, Lei},
  year = {2021},
  pages = {3024--3033},
  url = {https://openaccess.thecvf.com/content/CVPR2021/html/Wang_Dense_Contrastive_Learning_for_Self-Supervised_Visual_Pre-Training_CVPR_2021_paper.html},
  urldate = {2024-10-30},
  booktitle = {Proceedings of the {{IEEE}}/{{CVF Conference}} on {{Computer Vision}} and {{Pattern Recognition}}},
  langid = {english}
}

@inproceedings{heMaskedAutoencodersAre2022,
  title = {Masked {{Autoencoders Are Scalable Vision Learners}}},
  author = {He, Kaiming and Chen, Xinlei and Xie, Saining and Li, Yanghao and Dollár, Piotr and Girshick, Ross},
  year = {2022},
  pages = {16000--16009},
  url = {https://openaccess.thecvf.com/content/CVPR2022/html/He_Masked_Autoencoders_Are_Scalable_Vision_Learners_CVPR_2022_paper.html},
  urldate = {2023-06-08},
  booktitle = {Proceedings of the {{IEEE}}/{{CVF Conference}} on {{Computer Vision}} and {{Pattern Recognition}}},
  langid = {english}
}

@inproceedings{taoSiameseImageModeling2023,
  title = {Siamese {{Image Modeling}} for {{Self-Supervised Vision Representation Learning}}},
  booktitle = {Proceedings of the {{IEEE}}/{{CVF Conference}} on {{Computer Vision}} and {{Pattern Recognition}}},
  author = {Tao, Chenxin and Zhu, Xizhou and Su, Weijie and Huang, Gao and Li, Bin and Zhou, Jie and Qiao, Yu and Wang, Xiaogang and Dai, Jifeng},
  year = {2023},
  pages = {2132--2141},
  urldate = {2024-11-08},
  langid = {english}
}

@inproceedings{baoBEiTBERTPreTraining2021,
  title = {{{BEiT}}: {{BERT Pre-Training}} of {{Image Transformers}}},
  shorttitle = {{{BEiT}}},
  author = {Bao, Hangbo and Dong, Li and Piao, Songhao and Wei, Furu},
  year = {2021},
  url = {https://openreview.net/forum?id=p-BhZSz59o4},
  urldate = {2024-08-30},
  booktitle = {International {{Conference}} on {{Learning Representations}}},
  langid = {english}
}

@inproceedings{liuMixMAEMixedMasked2023,
  title = {{{MixMAE}}: {{Mixed}} and {{Masked Autoencoder}} for {{Efficient Pretraining}} of {{Hierarchical Vision Transformers}}},
  shorttitle = {{{MixMAE}}},
  author = {Liu, Jihao and Huang, Xin and Zheng, Jinliang and Liu, Yu and Li, Hongsheng},
  year = {2023},
  pages = {6252--6261},
  url = {https://openaccess.thecvf.com/content/CVPR2023/html/Liu_MixMAE_Mixed_and_Masked_Autoencoder_for_Efficient_Pretraining_of_Hierarchical_CVPR_2023_paper.html},
  urldate = {2024-09-13},
  booktitle = {Proceedings of the {{IEEE}}/{{CVF Conference}} on {{Computer Vision}} and {{Pattern Recognition}}},
  langid = {english}
}

@inproceedings{kimProxyAnchorLoss2020,
  title = {Proxy {{Anchor Loss}} for {{Deep Metric Learning}}},
  author = {Kim, Sungyeon and Kim, Dongwon and Cho, Minsu and Kwak, Suha},
  year = {2020},
  pages = {3238--3247},
  urldate = {2024-05-08},
  booktitle = {Proceedings of the {{IEEE}}/{{CVF Conference}} on {{Computer Vision}} and {{Pattern Recognition}}}
}

@online{winkensContrastiveTrainingImproved2020,
  title = {Contrastive {{Training}} for {{Improved Out-of-Distribution Detection}}},
  author = {Winkens, Jim and Bunel, Rudy and Roy, Abhijit Guha and Stanforth, Robert and Natarajan, Vivek and Ledsam, Joseph R. and MacWilliams, Patricia and Kohli, Pushmeet and Karthikesalingam, Alan and Kohl, Simon and Cemgil, Taylan and Eslami, S. M. Ali and Ronneberger, Olaf},
  date = {2020-07-10},
  eprint = {2007.05566},
  eprinttype = {arxiv},
  doi = {10.48550/arXiv.2007.05566},
  url = {http://arxiv.org/abs/2007.05566},
  urldate = {2024-05-08},
  pubstate = {preprint}
}

@inproceedings{sehwagSSDUnifiedFramework2020,
  title = {{{SSD}}: {{A Unified Framework}} for {{Self-Supervised Outlier Detection}}},
  shorttitle = {{{SSD}}},
  author = {Sehwag, Vikash and Chiang, Mung and Mittal, Prateek},
  year = {2021},
  url = {https://openreview.net/forum?id=v5gjXpmR8J},
  urldate = {2024-05-08},
  booktitle = {International {{Conference}} on {{Learning Representations}}},
  langid = {english}
}

@online{mingHowExploitHyperspherical2023,
  title = {How to Exploit Hyperspherical Embeddings for Out-of-Distribution Detection?},
  author = {Ming, Yifei and Sun, Yiyou and Dia, Ousmane and Li, Yixuan},
  year = {2023},
  eprint = {2203.04450},
  eprinttype = {arxiv},
  url = {http://arxiv.org/abs/2203.04450},
  urldate = {2023-03-24},
  pubstate = {preprint}
}

@article{kaplan2020scaling,
  title={Scaling Laws for Neural Language Models},
  author={Kaplan, Jared and McCandlish, Sam and Henighan, Tom and Brown, Tom B and Chess, Benjamin and Child, Rewon and Gray, Scott and Radford, Alec and Wu, Jeffrey and Amodei, Dario},
  journal={arXiv preprint arXiv:2001.08361},
  year={2020}
}

@inproceedings{hendrycksNaturalAdversarialExamples2021,
  title = {Natural {{Adversarial Examples}}},
  author = {Hendrycks, Dan and Zhao, Kevin and Basart, Steven and Steinhardt, Jacob and Song, Dawn},
  year = {2021},
  pages = {15262--15271},
  url = {https://openaccess.thecvf.com/content/CVPR2021/html/Hendrycks_Natural_Adversarial_Examples_CVPR_2021_paper.html},
  urldate = {2024-11-21},
  booktitle = {Proceedings of the {{IEEE}}/{{CVF Conference}} on {{Computer Vision}} and {{Pattern Recognition}}},
  langid = {english}
}

@inproceedings{hendrycksManyFacesRobustness2021,
  title = {The {{Many Faces}} of {{Robustness}}: {{A Critical Analysis}} of {{Out-of-Distribution Generalization}}},
  shorttitle = {The {{Many Faces}} of {{Robustness}}},
  author = {Hendrycks, Dan and Basart, Steven and Mu, Norman and Kadavath, Saurav and Wang, Frank and Dorundo, Evan and Desai, Rahul and Zhu, Tyler and Parajuli, Samyak and Guo, Mike and Song, Dawn and Steinhardt, Jacob and Gilmer, Justin},
  year = {2021},
  pages = {8340--8349},
  url = {https://openaccess.thecvf.com/content/ICCV2021/html/Hendrycks_The_Many_Faces_of_Robustness_A_Critical_Analysis_of_Out-of-Distribution_ICCV_2021_paper.html},
  urldate = {2024-11-21},
  booktitle = {Proceedings of the {{IEEE}}/{{CVF International Conference}} on {{Computer Vision}}},
  langid = {english}
}

@inproceedings{wangLearningRobustGlobal2019,
  title = {Learning {{Robust Global Representations}} by {{Penalizing Local Predictive Power}}},
  booktitle = {Advances in {{Neural Information Processing Systems}}},
  author = {Wang, Haohan and Ge, Songwei and Lipton, Zachary and Xing, Eric P},
  year = {2019},
  volume = {32},
  publisher = {Curran Associates, Inc.},
  url = {https://proceedings.neurips.cc/paper/2019/hash/3eefceb8087e964f89c2d59e8a249915-Abstract.html},
  urldate = {2024-11-20}
}

@inproceedings{netzer2011reading,
  title={Reading digits in natural images with unsupervised feature learning},
  author={Netzer, Yuval and Wang, Tao and Coates, Adam and Bissacco, Alessandro and Wu, Baolin and Ng, Andrew Y and others},
  booktitle={NIPS workshop on deep learning and unsupervised feature learning},
  volume={2011},
  number={5},
  pages={7},
  year={2011},
  organization={Granada}
}

@article{yu2015lsun,
  title={Lsun: Construction of a large-scale image dataset using deep learning with humans in the loop},
  author={Yu, Fisher and Seff, Ari and Zhang, Yinda and Song, Shuran and Funkhouser, Thomas and Xiao, Jianxiong},
  journal={arXiv preprint arXiv:1506.03365},
  year={2015}
}

@inproceedings{heMaskRCNN2017,
  title = {Mask {{R-CNN}}},
  author = {He, Kaiming and Gkioxari, Georgia and Dollar, Piotr and Girshick, Ross},
  date = {2017},
  pages = {2961--2969},
  url = {https://openaccess.thecvf.com/content_iccv_2017/html/He_Mask_R-CNN_ICCV_2017_paper.html},
  urldate = {2024-11-21},
  booktitle = {Proceedings of the {{IEEE International Conference}} on {{Computer Vision}}}
}

@article{papa2024survey,
  title={A survey on efficient vision transformers: algorithms, techniques, and performance benchmarking},
  author={Papa, Lorenzo and Russo, Paolo and Amerini, Irene and Zhou, Luping},
  journal={IEEE transactions on pattern analysis and machine intelligence},
  volume={46},
  number={12},
  pages={7682--7700},
  year={2024},
  publisher={IEEE}
}

@article{feng2023efficient,
  title={Efficient vision transformer via token merger},
  author={Feng, Zhanzhou and Zhang, Shiliang},
  journal={IEEE Transactions on Image Processing},
  volume={32},
  pages={4156--4169},
  year={2023},
  publisher={IEEE}
}

@article{mao2025prune,
  title={Prune and merge: Efficient token compression for vision transformer with spatial information preserved},
  author={Mao, Junzhu and Shen, Yang and Guo, Jinyang and Yao, Yazhou and Hua, Xiansheng and Shen, Hengtao},
  journal={IEEE Transactions on Multimedia},
  year={2025},
  publisher={IEEE}
}

@inproceedings{long2023beyond,
  title={Beyond attentive tokens: Incorporating token importance and diversity for efficient vision transformers},
  author={Long, Sifan and Zhao, Zhen and Pi, Jimin and Wang, Shengsheng and Wang, Jingdong},
  booktitle={Proceedings of the IEEE/CVF Conference on Computer Vision and Pattern Recognition},
  pages={10334--10343},
  year={2023}
}

@inproceedings{koner2024lookupvit,
  title={Lookupvit: Compressing visual information to a limited number of tokens},
  author={Koner, Rajat and Jain, Gagan and Jain, Prateek and Tresp, Volker and Paul, Sujoy},
  booktitle={European Conference on Computer Vision},
  pages={322--337},
  year={2024},
  organization={Springer}
}

@inproceedings{wei2023joint,
  title={Joint token pruning and squeezing towards more aggressive compression of vision transformers},
  author={Wei, Siyuan and Ye, Tianzhu and Zhang, Shen and Tang, Yao and Liang, Jiajun},
  booktitle={Proceedings of the IEEE/CVF conference on computer vision and pattern recognition},
  pages={2092--2101},
  year={2023}
}

@inproceedings{kim2024token,
  title={Token fusion: Bridging the gap between token pruning and token merging},
  author={Kim, Minchul and Gao, Shangqian and Hsu, Yen-Chang and Shen, Yilin and Jin, Hongxia},
  booktitle={Proceedings of the IEEE/CVF Winter Conference on Applications of Computer Vision},
  pages={1383--1392},
  year={2024}
}

@inproceedings{xu2022evo,
  title={Evo-vit: Slow-fast token evolution for dynamic vision transformer},
  author={Xu, Yifan and Zhang, Zhijie and Zhang, Mengdan and Sheng, Kekai and Li, Ke and Dong, Weiming and Zhang, Liqing and Xu, Changsheng and Sun, Xing},
  booktitle={Proceedings of the AAAI conference on artificial intelligence},
  volume={36},
  number={3},
  pages={2964--2972},
  year={2022}
}

@inproceedings{fayyaz2022adaptive,
  title={Adaptive token sampling for efficient vision transformers},
  author={Fayyaz, Mohsen and Koohpayegani, Soroush Abbasi and Jafari, Farnoush Rezaei and Sengupta, Sunando and Joze, Hamid Reza Vaezi and Sommerlade, Eric and Pirsiavash, Hamed and Gall, J{\"u}rgen},
  booktitle={European conference on computer vision},
  pages={396--414},
  year={2022},
  organization={Springer}
}

@InProceedings{Haurum_2023_ICCV,
    author    = {Haurum, Joakim Bruslund and Escalera, Sergio and Taylor, Graham W. and Moeslund, Thomas B.},
    title     = {Which Tokens to Use? Investigating Token Reduction in Vision Transformers},
    booktitle = {Proceedings of the IEEE/CVF International Conference on Computer Vision (ICCV) Workshops},
    month     = {October},
    year      = {2023},
    pages     = {773-783}
}

@inproceedings{lei2025rethinking,
  title={Rethinking Token Reduction with Parameter-Efficient Fine-Tuning in ViT for Pixel-Level Tasks},
  author={Lei, Cheng and Li, Ao and Yao, Hu and Zhu, Ce and Zhang, Le},
  booktitle={Proceedings of the Computer Vision and Pattern Recognition Conference},
  pages={14954--14964},
  year={2025}
}

@inproceedings{shi2024we,
  title={When do we not need larger vision models?},
  author={Shi, Baifeng and Wu, Ziyang and Mao, Maolin and Wang, Xin and Darrell, Trevor},
  booktitle={European Conference on Computer Vision},
  pages={444--462},
  year={2024},
  organization={Springer}
}

@inproceedings{zhai2022scaling,
  title={Scaling vision transformers},
  author={Zhai, Xiaohua and Kolesnikov, Alexander and Houlsby, Neil and Beyer, Lucas},
  booktitle={Proceedings of the IEEE/CVF conference on computer vision and pattern recognition},
  pages={12104--12113},
  year={2022}
}

@article{alabdulmohsin2023getting,
  title={Getting vit in shape: Scaling laws for compute-optimal model design},
  author={Alabdulmohsin, Ibrahim M and Zhai, Xiaohua and Kolesnikov, Alexander and Beyer, Lucas},
  journal={Advances in Neural Information Processing Systems},
  volume={36},
  pages={16406--16425},
  year={2023}
}

@inproceedings{cherti2023reproducible,
  title={Reproducible scaling laws for contrastive language-image learning},
  author={Cherti, Mehdi and Beaumont, Romain and Wightman, Ross and Wortsman, Mitchell and Ilharco, Gabriel and Gordon, Cade and Schuhmann, Christoph and Schmidt, Ludwig and Jitsev, Jenia},
  booktitle={Proceedings of the IEEE/CVF conference on computer vision and pattern recognition},
  pages={2818--2829},
  year={2023}
}

@inproceedings{xie2023data,
  title={On data scaling in masked image modeling},
  author={Xie, Zhenda and Zhang, Zheng and Cao, Yue and Lin, Yutong and Wei, Yixuan and Dai, Qi and Hu, Han},
  booktitle={Proceedings of the IEEE/CVF Conference on Computer Vision and Pattern Recognition},
  pages={10365--10374},
  year={2023}
}

@inproceedings{goyal2024scaling,
  title={Scaling Laws for Data Filtering--Data Curation cannot be Compute Agnostic},
  author={Goyal, Sachin and Maini, Pratyush and Lipton, Zachary C and Raghunathan, Aditi and Kolter, J Zico},
  booktitle={Proceedings of the IEEE/CVF Conference on Computer Vision and Pattern Recognition},
  pages={22702--22711},
  year={2024}
}

@inproceedings{wangscaling,
  title={Scaling Laws in Patchification: An Image Is Worth 50,176 Tokens And More},
  author={Wang, Feng and Yu, Yaodong and Shao, Wei and Zhou, Yuyin and Yuille, Alan and Xie, Cihang},
  booktitle={Forty-second International Conference on Machine Learning},
  year = {2025}
}

@inproceedings{ravishankar2025scaling,
  title={Scaling properties of diffusion models for perceptual tasks},
  author={Ravishankar, Rahul and Patel, Zeeshan and Rajasegaran, Jathushan and Malik, Jitendra},
  booktitle={Proceedings of the Computer Vision and Pattern Recognition Conference},
  pages={12945--12954},
  year={2025}
}

@inproceedings{yin2025towards,
  title={Towards precise scaling laws for video diffusion transformers},
  author={Yin, Yuanyang and Zhao, Yaqi and Zheng, Mingwu and Lin, Ke and Ou, Jiarong and Chen, Rui and Huang, Victor Shea-Jay and Wang, Jiahao and Tao, Xin and Wan, Pengfei and others},
  booktitle={Proceedings of the Computer Vision and Pattern Recognition Conference},
  pages={18155--18165},
  year={2025}
}

@article{dosovitskiy2020image,
  title={An image is worth 16x16 words: Transformers for image recognition at scale},
  author={Dosovitskiy, Alexey},
  journal={arXiv preprint arXiv:2010.11929},
  year={2020}
}

@inproceedings{yangMaskedImageContrastive2024,
  title={One Leaf Reveals the Season: Occlusion-Based Contrastive Learning with Semantic-Aware Views for Efficient Visual Representation},
  author={Xiaoyu Yang and Lijian Xu and Hongsheng Li and Shaoting Zhang},
  booktitle={Forty-second International Conference on Machine Learning},
  year={2025},
  url={https://openreview.net/forum?id=toZOqONu9x}
}

@inproceedings{yang2025adapting,
  author = {Yang, Xiaoyu and Lu, Jie and Yu, En},
  booktitle={The Thirteenth International Conference on Learning Representations},
  editor = {Y. Yue and A. Garg and N. Peng and F. Sha and R. Yu},
  pages = {90869--90891},
  title = {Adapting Multi-modal Large Language Model to Concept Drift From Pre-training Onwards},
  url = {https://proceedings.iclr.cc/paper_files/paper/2025/file/e25d87b8a42ee3f0d5b3ef741ca13031-Paper-Conference.pdf},
  volume = {2025},
  year = {2025}
}

@article{yu2025drift,
  title={Drift-aware collaborative assistance mixture of experts for heterogeneous multistream learning},
  author={Yu, En and Lu, Jie and Wang, Kun and Yang, Xiaoyu and Zhang, Guangquan},
  journal={arXiv preprint arXiv:2508.01598},
  year={2025}
}

@article{yang2025walking,
  title={Walking the tightrope: Disentangling beneficial and detrimental drifts in non-stationary custom-tuning},
  author={Yang, Xiaoyu and Lu, Jie and Yu, En},
  journal={arXiv preprint arXiv:2505.13081},
  year={2025}
}

@article{yang2024adaptingmultimodallargelanguage,
  title={Adapting Multi-modal Large Language Model to Concept Drift From Pre-training Onwards}, 
  author={Yang, Xiaoyu and Lu, Jie and Yu, En},
  journal={arXiv preprint arXiv:2405.13459},
  year={2024}
}

@article{chen2024general,
  title={A general variation-driven network for medical image synthesis},
  author={Chen, Yufei and Yang, Xiaoyu and Yue, Xiaodong and Lin, Xiang and Zhang, Qi and Fujita, Hamido},
  journal={Applied Intelligence},
  volume={54},
  number={4},
  pages={3295--3307},
  year={2024},
  publisher={Springer Nature BV}
}

@article{yang2022local,
  title={Local linear embedding based interpolation neural network in pancreatic tumor segmentation},
  author={Yang, Xiaoyu and Chen, Yufei and Yue, Xiaodong and Ma, Chao and Yang, Panpan},
  journal={Applied Intelligence},
  volume={52},
  number={8},
  pages={8746--8756},
  year={2022},
  publisher={Springer}
}

@article{yang2024segmentation,
  title={Segmentation and vascular vectorization for coronary artery by geometry-based cascaded neural network},
  author={Yang, Xiaoyu and Xu, Lijian and Yu, Simon and Xia, Qing and Li, Hongsheng and Zhang, Shaoting},
  journal={IEEE Transactions on Medical Imaging},
  year={2024},
  publisher={IEEE}
}

@article{yangTdistributedSphericalFeature2023,
  title = {T-Distributed {{Spherical Feature Representation}} for {{Imbalanced Classification}}},
  author = {Yang, Xiaoyu and Chen, Yufei and Yue, Xiaodong and Xu, Shaoxun and Ma, Chao},
  year = {2023},
  journal = {Proceedings of the AAAI Conference on Artificial Intelligence},
  volume = {37},
  year = {2023},
  number = {9},
  pages = {10825--10833},
  issn = {2374-3468},
  doi = {10.1609/aaai.v37i9.26284},
  url = {https://ojs.aaai.org/index.php/AAAI/article/view/26284},
  urldate = {2024-03-06},
  issue = {9},
  langid = {english}
}

@article{yang2024enhancing,
  title={Enhancing Visual Grounding and Generalization: A Multi-Task Cycle Training Approach for Vision-Language Models}, 
  author={Xiaoyu Yang and Lijian Xu and Hao Sun and Hongsheng Li and Shaoting Zhang},
  year={2024},
  journal={arXiv preprint arXiv:2311.12327},
}

@article{yang2025learning,
  title={Learning from All: Concept Alignment for Autonomous Distillation from Multiple Drifting MLLMs},
  author={Yang, Xiaoyu and Lu, Jie and Yu, En},
  journal={arXiv preprint arXiv:2510.04142},
  year={2025}
}

@article{yang2025causal,
  title={Causal-Informed Contrastive Learning: Towards Bias-Resilient Pre-training under Concept Drift},
  author={Yang, Xiaoyu and Lu, Jie and Yu, En},
  journal={arXiv e-prints},
  pages={arXiv--2502},
  year={2025}
}

@inproceedings{yang2021variational,
  title={Variational synthesis network for generating micro computed tomography from cone beam computed tomography},
  author={Yang, Xiaoyu and Chen, Yufei and Yue, Xiaodong and Lin, Xiang and Zhang, Qi},
  booktitle={2021 IEEE International Conference on Bioinformatics and Biomedicine (BIBM)},
  pages={1611--1614},
  year={2021},
  organization={IEEE}
}
}

\clearpage
\setcounter{page}{1}
\maketitlesupplementary

\section{Related Works}
\label{appendix:relatedwork}

\subsection{Scaling Laws in Vision Models}

A growing body of work has explored scaling behaviors in modern vision systems. Early studies on Vision Transformers reveal consistent power-law relationships between model size, data scale, and compute. Works such as SVT \cite{zhai2022scaling} and GVS \cite{alabdulmohsin2023getting} demonstrate that optimal performance emerges only when depth, width, data size, and compute are jointly scaled, and that well-structured scaling of model shape can outperform naïve parameter growth. Similarly, SL-CLIP \cite{cherti2023reproducible} extends these findings to contrastive language–image pre-training, showing stable scaling exponents across architectures and dataset variations.

Beyond model parameters, several works investigate how data scaling affects representation quality. SimMIM \cite{xie2023data} shows that masked image modeling (MIM) yields predictable scaling with increasing dataset size, while SLDF \cite{goyal2024scaling} argues that data curation quality itself follows compute-dependent scaling behavior. These studies highlight the importance of scaling both quantity and quality of visual data, but operate under fixed tokenization schemes.

A complementary direction examines input token scaling. Scaling Laws in Patchification \cite{wangscaling} demonstrates that reducing patch size—and thus dramatically increasing the number of tokens—consistently improves performance. In contrast, $S^{2}$\cite{shi2024we} shows that multi-scale inference can substitute for excessively large models, suggesting a coupling between input resolution and model capacity. Yet these works treat tokens as fixed spatial units, without formalizing their semantic redundancy.

In parallel, recent work on concept drift and non-stationary data distributions provides another scaling perspective that is highly relevant to real-world large-scale training. Yang et al. investigate how visual and multimodal models behave when both data volume and data distribution evolve over time, revealing drift-dependent scaling dynamics during large-scale pre-training \cite{yang2024adaptingmultimodallargelanguage, yang2025walking, yang2025causal}. Similarly, multi-stream and multi-teacher learning scenarios exhibit heterogeneous scaling behaviors under drift \cite{yu2025drift, yang2025learning}, further emphasizing that scaling laws should account not only for static datasets but also for distributional evolution. These findings align with our motivation to study scaling from the perspective of semantic sufficiency rather than raw token count.

Scaling analyses have also been extended to generative models, including image and video diffusion transformers. Recent studies \cite{ravishankar2025scaling,yin2025towards} show that diffusion architectures follow predictable trends with respect to model size, training compute, and sequence length, mirroring patterns in language and vision transformers.

In contrast to prior work, which primarily scales models, data, or resolution, our approach investigates scaling from the image perspective. By modeling visual tokens through the MDL lens and compressing them into semantic bases via Orthogonal Filtering, we uncover The Law of Parametric Efficiency Priority: larger vision models require fewer tokens to achieve full semantic reconstruction. This shifts the focus from scaling model capacity to scaling the semantic sufficiency of visual tokens, offering a new dimension to efficient vision model scaling.

\subsection{Efficient Token in Vision Models}

Efficient token utilization in Vision Transformers has been extensively explored as a primary route to reduce the quadratic cost of self-attention. A recent survey \cite{papa2024survey} on efficient ViTs systematically categorizes these efforts into compact architectures, pruning, distillation, and quantization, and highlights token-level compression as one of the most effective levers for controlling FLOPs while preserving accuracy. Our work falls into this token-centric line, but differs by grounding token reduction in an MDL-based semantic perspective and by explicitly linking the optimal token budget to model capacity via a scaling law.

The mainstream focuses on token merging as a way to compress spatially or semantically redundant patches into fewer tokens. TMViT \cite{feng2023efficient} introduces a similarity-based Token Merger that aggregates redundant tokens throughout the network, achieving substantial FLOP reductions with minor accuracy loss on classification benchmarks. 
Moreover, PM-ViT\cite{mao2025prune} extends this idea by pruning redundant tokens and then merging neighboring ones to better preserve spatial structure, showing benefits not only for classification but also for dense prediction tasks where spatial consistency matters. 
Furthermore, BAT \cite{long2023beyond} further emphasizes that simply keeping attentive tokens is insufficient, and proposes a decoupling–merging strategy that jointly accounts for token importance and global diversity, improving the trade-off between complexity and accuracy by preserving diverse global semantics while merging similar inattentive tokens. 
Meanwhile, DR-PEFT \cite{lei2025rethinking} investigates how token reduction interacts with PEFT schemes in segmentation and other pixel-level tasks, showing that naive token pruning can severely harm fine-grained predictions and proposing PEFT designs that are robust to token reduction.
Besides, LookupViT \cite{koner2024lookupvit} pushes merging to an extreme by compressing high-resolution features into a fixed, small set of “lookup” tokens that are processed by heavier blocks while the original tokens go through lightweight layers, yielding up to 2×FLOP reduction with strong robustness and generalization. 
Additionally, another line of work combines pruning and fusing to mitigate the information loss of hard token dropping, such as TPS \cite{wei2023joint} and Token Fusion \cite{kim2024token}. 
Compared to these heuristically designed merging rules, our Orthogonal Filtering module explicitly constructs a compact set of semantic bases and uses MDL to interpret merging as learning a minimal basis for the visual semantic space.

\begin{table*}[htbp]
\centering
\setlength{\tabcolsep}{3mm}{
\begin{tabular}{@{}cccccc@{}}
\toprule
          & \begin{tabular}[c]{@{}c@{}}Hidden\\      Dimensions\end{tabular} & \begin{tabular}[c]{@{}c@{}}Encoder\\      Layers\end{tabular} & \begin{tabular}[c]{@{}c@{}}Attention\\      Heads\end{tabular} & \begin{tabular}[c]{@{}c@{}}FFN\\      Intermediate\end{tabular} & Parameters \\ \midrule
ViT-Tiny  & 192                                                              & 12                                                            & 3                                                              & 768                                                             & 5.7M       \\
ViT-Small & 384                                                              & 12                                                            & 6                                                              & 1536                                                            & 22.1M      \\
ViT-Base  & 768                                                              & 12                                                            & 12                                                             & 3072                                                            & 86M        \\
ViT-Large & 1024                                                             & 24                                                            & 16                                                             & 4096                                                            & 304M       \\
ViT-Huge  & 1280                                                             & 32                                                            & 16                                                             & 5120                                                            & 632M       \\ \bottomrule
\end{tabular}
}
\caption{Parameters of ViT Backbones.}
\label{tab:vits}
\end{table*}

Dynamic or input-adaptive token selection is another major direction. Evo-ViT \cite{xu2022evo} proposes a slow-fast token evolution mechanism that progressively selects informative tokens using class-token attention and processes informative and uninformative tokens through different computation paths, enabling dynamic acceleration from the beginning of training. 
Besides, Adaptive Token Sampling (ATS) \cite{fayyaz2022adaptive} introduces a differentiable, parameter-free module that scores tokens and adaptively samples a variable number of tokens per image, achieving up to 2× GFLOP reduction on both image and video classification tasks without retraining in its plug-and-play mode. 
Moreover, a systematic empirical study of token reduction strategies is provided \cite{Haurum_2023_ICCV}, showing that the choice of which tokens to retain has a strong impact on accuracy and robustness across tasks. 
Similarly, BAT \cite{long2023beyond} also highlights the importance of considering both token importance and diversity rather than naive attention thresholds. 
Our method is complementary: instead of focusing on the selection policy alone, we provide a principled framework that interprets selection and merging as searching for a minimal semantic code consistent with the model’s capacity.

Meanwhile, advancing efficient representations under distribution shift or concept drift provides an orthogonal but complementary perspective. A line of works by Yang et al. \cite{yangTdistributedSphericalFeature2023, yang2024enhancing} develops efficient, semantically grounded contrastive methods and multi-task visual generalization, while Yang et al. \cite{yang2022local, yang2021variational, chen2024general, yang2024segmentation} explore efficient and interpretable medical-image feature learning, aligning with the goal of compressing visual information while preserving task-critical semantics. These studies collectively reinforce the broader motivation to design compressed yet semantically faithful visual representations.

Overall, existing works on efficient token utilization mainly design heuristic criteria or architectural modules to decide which tokens to drop, merge, or dynamically process, and evaluate them empirically across tasks. Our approach situates token reduction within a unified MDL framework, interprets compressed tokens as orthogonal semantic bases of the visual semantic space, and, crucially, establishes a quantitative scaling relationship between model size and the minimal token budget. This perspective complements prior token pruning/merging and dynamic sampling methods, and provides a principled basis for designing future efficient vision transformers that scale to larger models with substantially fewer tokens.


\section{Building Visual Long Context Dataset:\\ PaperScope}

To support research on long-context visual modeling and scalable token utilization, we construct PaperScope, a large-scale dataset of high-resolution academic document images derived from peer-reviewed machine learning papers. PaperScope provides a unified testbed for evaluating visual models’ ability to process extremely long sequences, compress redundant visual tokens, and preserve high-level semantics under varying visible ratios. Below, we describe its statistics, characteristics, and potential uses.

PaperScope contains more than 17,000 long-form paper images, each averaging 15k–22k pixels in height and reaching up to 30k pixels in extreme cases. This yields tens of thousands of tokens per image under standard ViT patch sizes, making the dataset a natural stress test for long-sequence Transformers. Compared with page-level document datasets, PaperScope maintains the global continuity of academic layouts, including multi-column text, mathematical expressions, figures and plots, tables, and reference sections. This rich multimodal structure provides a realistic environment for examining how visual models process highly heterogeneous and semantically dense content.

\begin{table}[htbp]
\centering
\begin{tabular}{@{}lc@{}}
\toprule
Statistic                           & Value              \\ \midrule
Number   of Papers                  & 17,000+            \\
Avg.   pages per paper              & 9.8                \\
Avg.   stitched image height        & 15k–22k pixels     \\
Max   resolution observed           & ~1,200 × 30,000 px \\
Total   image volume                & 42.3 GB           \\
Average   tokens per image (ViT-16) & 50k–120k tokens    \\ \bottomrule
\end{tabular}
\caption{Statistics of PaperScope Dataset.}
\label{tab:my-table}
\end{table}

A distinctive characteristic of PaperScope is the substantial redundancy inherent in academic formatting. Similar textual layouts, repeated structural motifs, and homogeneous backgrounds create a setting where large portions of visual tokens contribute limited new semantic information. Such redundancy makes the dataset particularly well-suited for evaluating token merging, pruning, orthogonal basis learning, and other forms of semantic compression. These properties align naturally with our Orthogonal Filtering framework, enabling systematic investigation of how model capacity interacts with token reduction and how larger models achieve complete semantic reconstruction with fewer visible tokens.

Beyond reconstruction and token-efficiency tasks, PaperScope enables a wide range of long-context visual benchmarks. Tasks such as document-level semantic prediction, figure–caption association, section boundary identification, or layout-aware classification can all be performed at the full-paper level, requiring models to integrate information across extremely long spatial spans. The dataset also supports studying scaling laws between token visibility, model size, and semantic fidelity, providing empirical insights into how expanding model capacity mitigates the need for dense visual tokens—an observation central to the empirical regularities established in our work.

Finally, PaperScope is curated entirely from publicly accessible research papers and contains no private or sensitive information. Its construction adheres to conference usage policies, and its purpose is strictly academic: to advance research on efficient long-context visual modeling, image redundancy analysis, and scalable Transformer architectures. As far as we are aware, PaperScope represents one of the largest and most semantically coherent datasets of long-document images, offering an invaluable resource for exploring the limits of visual token compression and the training of next-generation large vision models.

\section{Implementation Details}

We adopt standard ViT backbones of different capacities, ranging from Tiny to Huge. Their architectural configurations, including hidden dimensions, depth, attention heads, and FFN intermediate sizes, are summarized in Table~\ref{tab:vits}. From the architectural perspective, ViT families follow a clear scaling trajectory—wider and deeper models consistently bring stronger representation capacity. As shown in Table~\ref{tab:vits}, this progression from Tiny to Huge forms the backbone foundation upon which we further study how model capacity interacts with token efficiency, revealing that larger models can sustain complete semantic reconstruction with substantially fewer tokens.

\begin{table}[htbp]
    \centering
    \setlength{\tabcolsep}{3mm}{
        \begin{tabular}{@{}lcc@{}}
        \toprule
                            & ViT-B/16    & ViT-L/16    \\ \midrule
        Training Epochs     & 100         & 100          \\
        Warmup Epochs       & 5           & 5           \\
        Optimizer           & AdamW       & AdamW       \\
        Base Learning Rate  & 1.5e-4      & 1.5e-4        \\
        Learning Rate Decay & Cosine      & Cosine      \\
        Adam $\beta$        & (0.9, 0.95) & (0.9, 0.95) \\
        Weight Decay        & 0.05        & 0.05        \\
        Batch Size          & 1,024       & 1,024       \\ \bottomrule
        \end{tabular}
        }  
    \caption{Examples of Fine-tuning Hyperparameters.}
    \label{table:finetune}        
\end{table}

We follow standard fine-tuning protocols for all ViT backbones \cite{dosovitskiy2020image}, as examples presented in Table \ref{table:finetune}. 
Each model is trained for 100 epochs with a 5-epoch warmup phase. We use AdamW as the optimizer, with a base learning rate of $1.5\times10^{-4}$, weight decay of 0.05, and $\beta_{1}= 0.9$, $\beta_{2}= 0.95$. The learning rate is scaled with the batch size and follows a cosine decay schedule throughout training. Besides, unless otherwise specified, all images are resized to a fixed resolution of 224×224. All models use the same optimization hyperparameters to ensure a fair comparison across model scales. And we execute our experiments on $8\times 2$ NVIDIA A100 GPUs.


\end{document}